# Holistic 3D Scene Parsing and Reconstruction from a Single RGB Image

Siyuan Huang[1,2], Siyuan Qi[1,2], Yixin Zhu[1,2],
Yinxue Xiao[1], Yuanlu Xu[1,2], and Song-Chun Zhu[1,2]

[1] University of California, Los Angeles
[2] International Center for AI and Robot Autonomy (CARA)

**Abstract.** We propose a computational framework to jointly parse a single RGB image and reconstruct a holistic 3D configuration composed by a set of CAD models using a stochastic grammar model. Specifically, we introduce a Holistic Scene Grammar (HSG) to represent the 3D scene structure, which characterizes a joint distribution over the functional and geometric space of indoor scenes. The proposed HSG captures three essential and often latent dimensions of the indoor scenes: i) latent human context, describing the affordance and the functionality of a room arrangement, ii) geometric constraints over the scene configurations, and iii) physical constraints that guarantee physically plausible parsing and reconstruction. We solve this joint parsing and reconstruction problem in an analysis-by-synthesis fashion, seeking to minimize the differences between the input image and the rendered images generated by our 3D representation, over the space of depth, surface normal, and object segmentation map. The optimal configuration, represented by a parse graph, is inferred using Markov chain Monte Carlo (MCMC), which efficiently traverses through the non-differentiable solution space, jointly optimizing object localization, 3D layout, and hidden human context. Experimental results demonstrate that the proposed algorithm improves the generalization ability and significantly outperforms prior methods on 3D layout estimation, 3D object detection, and holistic scene understanding.

## 1 Introduction

The complexity and richness of human vision are not only reflected by the ability to recognize visible objects, but also to reason about the latent actionable information [1], including inferring latent human context as the functionality of a scene [2, 3], reconstructing 3D hierarchical geometric structure [4, 5], and complying with the physical constraints that guarantee the physically plausible scene configurations [6]. Such rich understandings of an indoor scene are the essence for building an intelligent computational system, which transcends the prevailing appearance- and geometry-based recognition tasks to account also for the deeper reasoning of observed images or patterns.

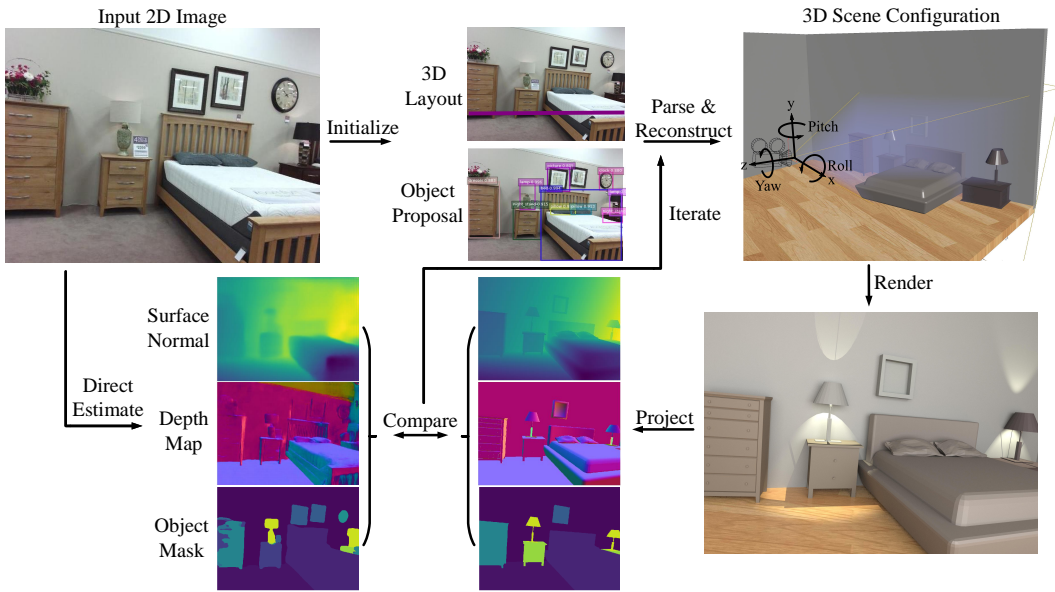

Fig. 1: Illustration of the proposed holistic 3D indoor scene parsing and reconstruction in an analysis-by synthesis fashion. A 3D representation is initialized by individual vision modules (*e.g.*, object detection, 2D layout estimation). A joint inference algorithm compares the differences between the rendered normal, depth, and segmentation map with the ones estimated directly from the input RGB image, and adjust the 3D structure iteratively.

One promising direction is *analysis-by-synthesis* [7] or "vision as inverse graphics" [8,9]. In this paradigm, computer vision is treated as an inverse problem as opposed to computer graphics, of which the goal is to reverse-engineer hidden factors occurred in the physical process that produces observed images.

In this paper, we embrace the concept of vision as inverse graphics, and propose a holistic 3D indoor scene parsing and reconstruction algorithm that simultaneously reconstructs the functional hierarchy and the 3D geometric structure of an indoor scene from a single RGB image. Figure 1 schematically illustrates the analysis-by-synthesis inference process. The joint inference algorithm takes proposals from various vision modules and infers the 3D structure by comparing various projections (*i.e.*, depth, normal, and segmentation) rendered from the recovered 3D structure with the ones directly estimated from an input image.

Specifically, we introduce a Holistic Scene Grammar (HSG) to represent the hierarchical structure of a scene. As illustrated in Figure 2, our HSG decomposes a scene into latent groups in the *functional space* (*i.e.*, hierarchical structure including activity groups) and object instances in the *geometric space* (*i.e.*, CAD models). For the functional space, in contrast to the conventional method that only models the object-object relations, we propose a novel method to model human-object relations by imagining latent human in activity groups to further help explain and parse the observed image. For the geometric space, the geometric attributes (*e.g.*, size, position, orientation) of individual objects are taken into considerations, as well as the geometric relations (*e.g.*, supporting relation) among them. In addition, physical constraints (*e.g.*, collision among the objects, violations of the layout) are incorporated to generate a physically plausible 3D parsing and reconstruction of the observed image.

Here, an indoor scene is represented by a parse graph (**pg**) of a grammar, which consists of a hierarchical structure and a Markov random field (MRF) over terminal nodes that captures the rich contextual relations between objects and room layout (*i.e.*, the room configuration of walls, floors, and ceilings).

A maximum a posteriori probability (MAP) estimate is designed to find the optimal solution that parses and reconstructs the observed image. The likelihood measures the similarity between the observed image and the rendered images projected from the inferred **pg** onto various 2D image spaces. Thus, the **pg** can be iteratively refined by sampling an MCMC with simulated annealing based on posterior probability. We evaluate our method on a large-scale RGB-D dataset by comparing the reconstructed 3D indoor rooms with the ground-truth.

### 1.1 Related Work

**Scene Parsing:** Existing scene parsing approaches fall into two streams. i) Discriminative approaches [10–16] classify each pixel to a semantic label. Although prior work has achieved high accuracy in labeling the pixels, these methods lack a general representation of visual vocabulary and a principle approach to exploring the semantic structure of a general scene. ii) Generative approaches [17–24] can distill scene structure, making it closer to human-interpretable structure of a scene, enabling potential applications in robotics, VQA, *etc.* In this paper, we combine those two streams in an analysis-by-synthesis framework to infer the hidden factors that generate the image.

**Scene Reconstruction from a Single Image:** Previous approaches [25–27] of indoor scene reconstruction from a single RGB image can be categorized into three streams. i) 2D or 3D room layout prediction by extracting geometric features and ranking the 3D cuboids proposals [28–35]. ii) By representing objects via geometric primitives or CAD models, previous approaches [36–44] utilize 3D object recognition or pose estimation to align object proposals to a RGB or depth image. iii) Joint estimation of the room layout and 3D objects with contexts [18, 19, 22–24, 33, 45, 46]. In particular, Izadinia *et al.* [33] show promising results in inferring the layout and objects without the contextual relations and physical constraints. In contrast, our method jointly models the hierarchical scene structure, hidden human context and physical constraints, providing a semantic representation for holistic scene understanding. Furthermore, the proposed method presents a joint inference algorithm using MCMC, which in theory can achieve a global optimal.

**Scene Grammar:** Scene grammar models have been used to infer the 3D structure and functionality from a RGB image [3, 17, 18, 47]. Our HSG differs from [17, 18] in two aspects: i) Our model represents the 3D objects with CAD models rather than geometric primitives, capable of modeling detail contextual relations (*e.g.*, supporting relation), which provides better realization of parsing and reconstruction. ii) We infer hidden human and activity groups in the HSG, which helps the explanation and parsing. Compared to [3, 47], we model and parse the 3D structure of objects and layouts from a single RGB image, rather than the labelled point-clouds using RGB-D images.

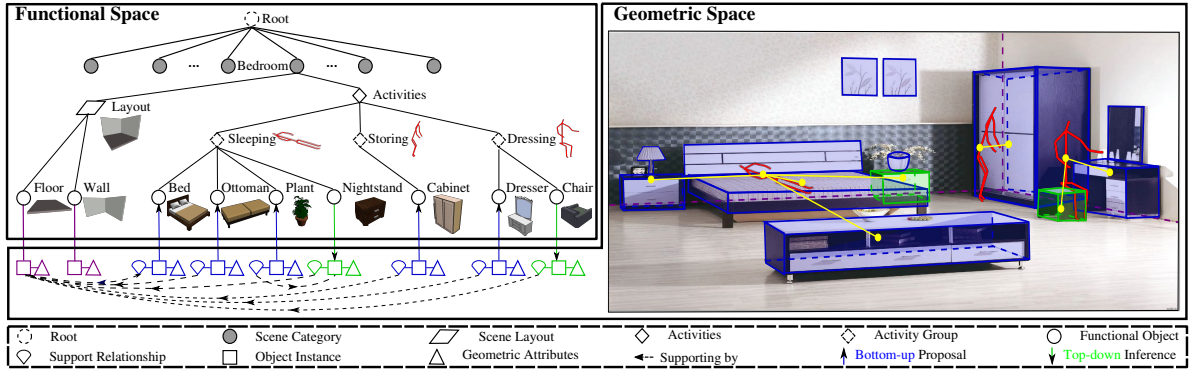

Fig. 2: An indoor scene represented by a parse graph (***pg***) of the HSG that spans across the functional space and the geometric space. The functional space characterizes the hierarchical structure and the geometric space encodes the spatial entities with contextual relations.

### 1.2   Contributions

This paper makes five major contributions:

1. We integrate geometry and physics to interpret and reconstruct indoor scenes with *CAD models*. We jointly optimize 3D room layouts and object configurations, largely improving the performance of scene parsing and reconstruction on SUN RGB-D dataset [45].

2. We incorporate hidden human context (*i.e.*, functionality) into our grammar, enabling to imagine *latent human pose* in each activity group by grouping and sampling. In this way, we can optimize the joint distribution of both visible and invisible [48] components of the scene.

3. We propose a complete computational framework to combine generative model (*i.e.*, a stochastic grammar), discriminative models (*i.e.*, direct estimations of depth, normal, and segmentation maps), and graphics engines (*i.e.*, rendered images) in scene parsing and reconstruction.

4. To the best of our knowledge, ours is the first work to use the inferred depth, surface normal and object segmentation map to assist parsing and reconstructing 3D scenes (both room layout and multiple objects). Note that [49] uses similar intermediate representation for a single object.

5. By learning the supporting relations among objects, the proposed method eliminates the widely adopted assumption in previous work that all objects must stand on the ground. Such flexibility of the model yields better parsing and reconstruction of the real-world scenes with complex object relations.

## 2    Holistic Scene Grammar

We represent the hierarchical structure of indoor scenes by a Holistic Scene Grammar (HSG). An HSG consists of a latent hierarchical structure in the functional space $\mathbb{F}$ and terminal object entities in the geometric space $\mathbb{G}$. The intuition is that, for man-made environments, the object arrangement in the geometric space should be a "projection" from the functional space (*i.e.*, human activities). The functional space as a probabilistic context free grammar (PCFG) captures the hierarchy of the functional groups, and the geometric space captures the spatial contexts among objects by defining an MRF on the terminal nodes. The two spaces together form a stochastic context-sensitive grammar (SCSG). The HSG starts from a root scene node and ends with a set of terminal nodes. An indoor scene is represented by a parse graph **pg** as illustrated in Figure 2.

*Definition:* The stochastic context-sensitive grammar HSG is defined as a 5-tuple $\langle S, V, R, E, P \rangle$. $S$ denotes the root node of the indoor scene. $V$ is the vertex set that includes both non-terminal nodes $V_f \in \mathbb{F}$ and terminal nodes $V_g \in \mathbb{G}$. $R$ denotes the production rule, and $E$ the contextual relations among the terminal nodes, which are represented by the horizontal links in the **pg**. $P$ is the probability model defined on the **pg**.

*Functional Space $\mathbb{F}$:* The non-terminal nodes $V_f = \{V_f^c, V_f^a, V_f^o, V_f^l\} \in \mathbb{F}$ consist of the scene category nodes $V_f^c$, activity group nodes $V_f^a$, objects nodes $V_f^o$, and layout nodes $V_f^l$.

*Geometric Space $\mathbb{G}$:* The terminal nodes $V_g = \{V_g^o, V_g^l\} \in \mathbb{G}$ are the CAD models of object entities and room layouts. Each object $v \in V_g^o$ is represented as a CAD model, and the object appearance is parameterized by its 3D size, location, and orientation. The room layout $v \in V_g^l$ is represented as a cuboid which is further decomposed into five planar surfaces of the room (left wall, right wall, middle wall, floor, and ceiling with respect to the camera coordinate).

*Production Rule $R$:* The following production rules are defined for HSG:
- $S \rightarrow V_f^c$: scene $\rightarrow$ category 1 | category 2 | ... (*e.g.*, scene $\rightarrow$ office | kitchen)
- $V_f^c \rightarrow V_f^a \cdot V_f^l$: category $\rightarrow$ activity groups $\cdot$ layout (*e.g.*, office $\rightarrow$ (walking, reading) $\cdot$ layout)
- $V_f^a \rightarrow V_f^o$: activity group $\rightarrow$ functional objects (*e.g.*, sitting $\rightarrow$ (desk, chair))

where $\cdot$ denotes the deterministic decomposition, | alternative explanations, and () combination. Contextual relations $E$ capture relations among objects, including their relative positions, relative orientations, grouping relations, and supporting relations. The objects could be supported by either other objects or the room layout; *e.g.*, a lamp could be supported by a night stand or the floor.

Finally, a scene configuration is represented by a **pg**, whose terminals are room layouts and objects with their attributes and relations. As shown in Figure 2, a **pg** can be decomposed as $\mathbf{pg} = (pg_f, pg_g)$, where $pg_f$ and $pg_g$ denote the functional part and geometric part of the **pg**, respectively. $E \in pg_g$ denotes the contextual relations in the terminal layer.

## 3    Probabilistic Formulation

The objective of the holistic scene parsing is to find an optimal **pg** that represents all the contents and relations observed in the scene. Given an input RGB image $I$, the optimal **pg** could be derived by an MAP estimator,

$$p(\mathbf{pg}|I) \propto p(\mathbf{pg}) \cdot p(I|\mathbf{pg}) \tag{1}$$

$$\propto p(pg_f) \cdot p(pg_g|pg_f) \cdot p(I|pg_g) \tag{2}$$

$$= \frac{1}{Z} \exp\left\{-\mathcal{E}(pg_f) - \mathcal{E}(pg_g|pg_f) - \mathcal{E}(I|pg_g)\right\}, \tag{3}$$

where the prior probability $p(\mathbf{pg})$ is decomposed into $p(pg_f)p(pg_g|pg_f)$, and $p(I|\mathbf{pg}) = p(I|pg_g)$ since the image space is independent of the functional space given the geometric space. We model the joint distribution with a Gibbs distribution; $\mathcal{E}(pg_f)$, $\mathcal{E}(pg_g|pg_f)$ and $\mathcal{E}(I|pg_g)$ are the corresponding energy terms.

**Functional Prior** $\mathcal{E}(pg_f)$ characterizes the prior of the functional aspect in a **pg**, which models the hierarchical structure and production rules in the functional space. For production rules of alternative explanations | and combination (), each rule selects child nodes and the probability of the selections is modeled with a multinomial distribution. The production rule · is deterministically expanded with probability 1. Given the production rules $R$, the energy can be written as $\mathcal{E}(pg_f) = \sum_{r_i \in R} -\log p(r_i)$.

**Geometric Prior** $\mathcal{E}(pg_g|pg_f)$ is the prior of the geometric aspect in a **pg**. Besides modeling the size, position and orientation distribution of each object, we also consider two types of contextual relations $E = \{E_s, E_a\}$ among the objects: i) relations $E_s$ between supported objects and their supporting objects; ii) relations $E_a$ between imagined human and objects in an activity group.

We define different potential functions for each type of contextual relations, constructing an MRF in the geometric space including four terms:

$$\mathcal{E}(pg_g|pg_f) = \mathcal{E}_{sc}(pg_g|pg_f) + \mathcal{E}_{spt}(pg_g|pg_f) + \mathcal{E}_{grp}(pg_g|pg_f) + \mathcal{E}_{phy}(pg_g). \tag{4}$$

• *Size Consistency* $\mathcal{E}_{sc}$ constrains the size of an object. $\mathcal{E}_{sc}(pg_g|pg_f) = \sum_{v_i \in V_g^o} -\log p(s_i|V_f^o)$, where $s_i$ denotes the size of object $v_i$. We model the distribution of object scale in a non-parametric way, *i.e.*, kernel density estimation (KDE).

• *Supporting Constraint* $\mathcal{E}_{spt}$ characterizes the contextual relations between supported objects and supporting objects (including floors, walls and ceilings). We model the distribution with their relative heights and overlapping areas:

$$\mathcal{E}_{spt}(pg_g|pg_f) = \sum_{(v_i, v_j) \in E_s} \mathcal{K}_o(v_i, v_j) + \mathcal{K}_h(v_i, v_j) - \lambda_s \log p\left(v_i, v_j \mid V_f^l, V_f^o\right), \tag{5}$$

where $\mathcal{K}_o(v_i, v_j) = 1 - area(v_i \cup v_j)/area(v_i)$ defines the overlapping ratio in xy-plane, and $\mathcal{K}_h(v_i, v_j)$ defines the relative height between the lower surface of $v_i$ and the upper surface of $v_j$. $\mathcal{K}_o(\cdot)$ and $\mathcal{K}_h(\cdot)$ is 0 if supporting object is floor and wall, respectively. $p(v_i, v_j|V_f^l, V_f^o)$ is the prior frequency of the supporting relation modeled by multinoulli distributions. $\lambda_s$ is a balancing constant.

• *Human-Centric Grouping Constraint* $\mathcal{E}_{grp}$. For each activity group, we imagine the invisible and latent human poses to help parse and understand the scene. The intuition is that the indoor scenes are designed to serve human daily activities, thus the indoor images should be jointly interpreted by the observed entities and the unobservable human activities. This is known as the *Dark Matter* [48] in computer vision that drives the visible components in the scene. Prior methods on scene parsing often merely model the object-object relations. In this paper, we go beyond passive observations to model the latent human-object relations, thereby proposing a human-centric grouping relationship and a joint inference algorithm over both the visible scene and the invisible latent human context. Specifically, for each activity group $v \in V_f^a$, we define correspondent imagined human with a six tuple $< y, \mu, t, r, s, \tilde{\mu} >$, where $y$ is the activity type, $\mu \in \mathbb{R}^{25 \times 3}$ is the mean human pose (represented by 25 joints) of activity type $y$, $t$ denotes the translation, $r$ denotes the rotation, $s$ denotes the scale, and $\tilde{\mu}$ is the imagined human skeleton: $\tilde{\mu} = \mu \cdot r \cdot s + t$. The energy among the imagined human and objects is defined as:

$$\mathcal{E}_{grp}(pg_g|pg_f) = \sum\nolimits_{v_i \in V_f^a} \mathcal{E}_{grp}(\tilde{\mu}_i|v_i)$$
$$= \sum\nolimits_{v_i \in V_f^a} \sum\nolimits_{v_j \in ch(v_i)} \mathcal{D}_d(\tilde{\mu}_i, \nu_j; \bar{d}) + \mathcal{D}_h(\tilde{\mu}_i, \nu_j; \bar{h}) + \mathcal{D}_o(\tilde{\mu}_i, \nu_j; \bar{o}), \quad (6)$$

where $ch(v_i)$ denotes the set of child nodes of $v_i$, $\nu_j$ denotes the 3D position of $v_j$. $\mathcal{D}_d(\cdot)$, $\mathcal{D}_h(\cdot)$ and $\mathcal{D}_o(\cdot)$ denote geometric distances, heights and orientation differences, respectively, calculated by the center of the imagined human pose to the object center subtracted by their mean (*i.e.*, $\bar{d}$, $\bar{h}$ and $\bar{o}$).

• *Physical Constraints:* Additionally, in order to avoid violating the physical laws during parsing, we define the physical constraints $\mathcal{E}_{phy}(pg_g)$ to penalize physical violations. Exceeding the room cuboid or overlapping among the objects are defined as violations. This term is formulated as:

$$\mathcal{E}_{phy}(pg_g) = \sum\nolimits_{v_i \in V_g^o}(\sum\nolimits_{v_j \in V_g^o \setminus v_i} \mathcal{O}_o(v_i, v_j) + \sum\nolimits_{v_j \in V_g^l} \mathcal{O}_l(v_i, v_j)), \quad (7)$$

where $\mathcal{O}_o(\cdot)$ denotes the overlapping area between objects, and $\mathcal{O}_l(\cdot)$ denotes the area of objects exceeding the layout.

**Likelihood** $\mathcal{E}(I|pg_g)$ characterizes the similarity between the observed image and the rendered image generated by the parsing results. Due to various lighting conditions, textures, and material properties, there will be an inevitable difference between the rendered RGB images and the observed scenes. Here, instead of using RGB images, we solve this problem in an *analysis-by-synthesis* fashion by comparing the depth, surface normal, and object segmentation map.

By combining generative models and discriminative models, the proposed approach tries to reverse-engineer the hidden factors that generate the observed image. Specifically, we first use discriminative methods to project the observed image $I$ to various feature spaces. In this paper, we directly estimate three intermediate images—depth map $\Phi_d(I)$, surface normal map $\Phi_n(I)$ and object segmentation map $\Phi_m(I)$, as the feature representation of the observed image $I$.

Meanwhile, a **pg** inferred by our method represents the 3D structure of the observed image, which is used to reconstruct image $I'$ to recover the correspond-ing depth map $\Phi_d(I')$, surface normal map $\Phi_n(I')$, and object segmentation map $\Phi_m(I')$ through a forward graphics rendering.

Finally, we compute the likelihood term by comparing these rendered results from the generative model with the directly estimated results calculated by the discriminative models. Specifically, the likelihood is computed by the pixel-wise differences between the two sets of maps,

$$\mathcal{E}(I|pg_g) = \mathcal{D}_p(\Phi_d(I), \Phi_d(I')) + \mathcal{D}_p(\Phi_n(I), \Phi_n(I')) + \mathcal{D}_p(\Phi_m(I), \Phi_m(I')), \quad (8)$$

where $\mathcal{D}_p(\cdot)$ is the sum of pixel-wise Euclidean distances between two maps. Note a weight is associated with each energy term, which is learned by cross-validation or set empirically.

## 4    Inference

Given a single RGB image as the input, the goal of inference is to find the optimal **pg** that best explains the hidden factors that generate the observed image while recovering the 3D scene structure. The inference includes three major steps:

- *Room geometry estimation:* estimate the room geometry by predicting the 2D room layout and the camera parameter, and by projecting the estimated 2D layout to 3D. Details are provided in subsection 4.1.

- *Objects initialization:* detect objects and retrieve CAD models correspondingly with the most similar appearance, then roughly estimate their 3D poses, positions, sizes, and initialize the support relations. See subsection 4.2.

- *Joint inference:* optimize the objects, layout and hidden human context in the 3D scene in an analysis-by-synthesis fashion by maximizing the posterior probability of the **pg**. Details are provided in subsection 4.3.

### 4.1    Room Geometry Estimation

Although recent approaches [33–35] are capable of generating a relatively robust prediction of the 2D room layout using CNN features, 3D room layout estimations are still inaccurate due to its sensitivity to camera parameter estimation in clustered scenes. To address the inconsistency between the 2D layout estimation and camera parameter estimation, we design a deep neural network to estimate the 2D layout, and use the layout heatmap to estimate the camera parameter.

**2D Layout Estimation:** Similar to [34], we represent the 2D layout with its room layout type and keypoint positions. The network structure is provided in the *supplementary material*. The network optimizes the Euclidean loss for layout heatmap regression and the cross-entropy loss for room type estimation.

**Camera Parameter:** Traditional geometry-based method [28] computes the camera parameter by estimating the vanishing points from the observed image, which is sensitive and unstable in cluttered indoor scenes with heavy occlusions. Inspired by [43], we propose a learning-based method that uses the keypoints heatmaps to predict the camera parameters, *i.e.*, focal length, together with the yaw, pitch, and roll angles of the camera. Since the yaw angle has already been incorporated into the evaluation of room layout, we estimate the remaining three variables (focal length, pitch and roll) by stacking four FC layers (1024-128-16-3) on the keypoint heatmaps.

**3D Layout Initialization:** Using the estimated 2D layout and camera parameters, we project the corners of the 2D layout to 3D in order to obtain a 3D room cuboid. We assume the cameras and the ceilings are $1.2m$ and $3.0m$ high, respectively. For simplicity, we translate and rotate the 3D rooms so that one of the visible room corners is at the origin of the world coordinate system.

### 4.2 Objects Initialization

We fine-tune the Deformable Convolutional Networks [50] using Soft-NMS [51] to detect 2D bounding boxes. To initialize the 3D objects, we retrieve the most similar CAD models and initialize their 3D poses, sizes, and positions.

**Model Retrieval:** We consider all the models in the ShapeNetSem repository [52, 53] and render each model from 48 viewpoints consisting of uniformly sampled 16 azimuth and 3 elevation angles. We extract $7 \times 7$ features from the ROI-pooling layer of the fine-tuned detector of images in the detected bounding boxes and candidate rendered images. By ranking the cosine distance between each detected object feature and rendered image feature in the same object category, we obtain the top-10 CAD models with corresponding poses.

**Geometric Attributes Estimation:** The geometric attributes of an object are represented by a 9D vector of 3D pose, position, and size, where 3D poses are initialized from the retrieval procedure. Prior work roughly projected 2D points to 3D, and recovered the 3D position and size by assuming that all the objects are on the floor. Such approach shows limitations in complex scenarios.

Without making the above assumption, we estimate the depth of each object by computing the average depth value of the pixels that are in both the detection bounding box and the segmentation map. We then compute its 3D position using the depth value. Empirically, this approach is more robust since per-pixel depth estimation error is small even in cluttered scenes. To avoid the alignment problem of the 2D bounding boxes, we initialize the object size by sampling object sizes from a learned distribution and choose the one with the largest probability.

**Supporting Relation Estimation:** For each object $v_i \in V_f^o$, we find its supporting object $v_j^*$ of minimal supporting energy from objects or layout:

$$v_j^* = \operatorname*{arg\,min}_{v_j} \mathcal{K}_o(v_i, v_j) + \mathcal{K}_h(v_i, v_j) - \lambda_s \log p(v_i, v_j | V_f^l, V_f^o), \quad v_j \in (V_f^l, V_f^o). \quad (9)$$

### 4.3 Joint Inference

Given an image $I$, we first estimate the room geometry, object attributes and relations as described in the above two subsections. As summarized in Alg.1, the joint inference includes: (1) optimize the objects and layout (Figure 3); (2) group objects, assign activity label and imagine human pose in each activity group; and (3) optimize the objects, layout and human pose iteratively.

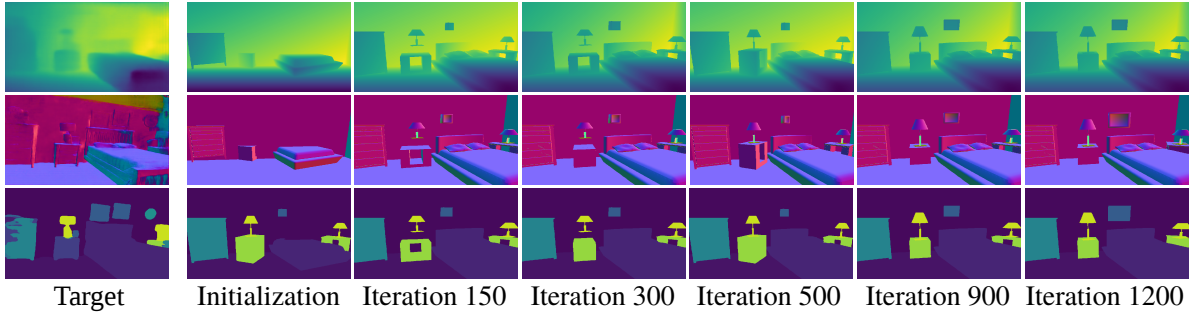

| Target | Initialization | Iteration 150 | Iteration 300 | Iteration 500 | Iteration 900 | Iteration 1200 |

Fig. 3: The process of joint inference of objects and layout by MCMC with simulated annealing. **Top**: depth maps. **Middle**: normal maps. **Bottom**: object segmentation maps. Objects and layout are optimized iteratively.

In each step, we use distinct MCMC processes. Specifically, to traverse non-differentiable solution spaces, we design Markov chain dynamics $\{q_1^o, q_2^o, q_3^o\}$ for objects, $\{q_1^l, q_2^l\}$ for layout, and $\{q_1^h, q_2^h, q_3^h\}$ for human pose. Specifically,

• *Object Dynamics:* Dynamics $q_1^o$ adjusts the position of a random object, which translates the object center in one of the three Cartesian coordinate axes. Instead of translating the object center and changing the object size directly, Dynamics $q_2^o$ translates one of the six faces of the cuboid to generate a smoother diffusion. Dynamics $q_3^o$ proposes rotation of the object with a specified angle. Each dynamic can diffuse in two directions, *e.g.*, each object can translate in direction of '$+x$' and '$-x$', or rotate in direction of clockwise and counterclockwise. By computing the local gradient of $P(\mathbf{pg}|I)$, the dynamics propose to move following the direction of the gradient with a proposal probability of 0.8, or the inverse direction of the gradient with proposal probability of 0.2.

• *Layout Dynamics:* Dynamics $q_1^l$ translates the faces of the layout, which also optimizes the camera height when translating the floor. Dynamics $q_2^l$ rotates the layout.

• *Human pose Dynamics* $q_1^h$, $q_2^h$ and $q_3^h$ are designed to translate, rotate and scale the human pose, respectively.

Given the current $\mathbf{pg}$, each dynamic will propose a new $\mathbf{pg}'$ according to a proposal probability $p(\mathbf{pg}'|\mathbf{pg}, I)$. The proposal is accepted according to an acceptance probability $\alpha(\mathbf{pg} \rightarrow \mathbf{pg}')$ defined by the Metropolis-Hasting algorithm [54]:

$$\alpha(\mathbf{pg} \rightarrow \mathbf{pg}') = \min(1, \frac{p(\mathbf{pg}|\mathbf{pg}', I)p(\mathbf{pg}'|I)}{p(\mathbf{pg}'|\mathbf{pg}, I)p(\mathbf{pg}|I)}). \tag{10}$$

In step (2), we group objects and assign activity labels. For each type of activity, there is a object category which has the highest occurrence frequency (*i.e.*, chair in activity 'reading'). Intuitively, the correspondence between objects and activities should be n-to-n but not n-to-one, which means each object can belong to several activity groups. In order to find out all possible activity groups, for each type of activity, we define an activity group around each major object and incorporate nearby objects (within a distance threshold) with prior larger

---

**Algorithm 1** Joint inference algorithm

---

1: **Given** Image $I$, initialized parse graph $\mathbf{pg}_{\text{init}}$
2: **procedure** STEP1($V_g^o, V_g^l$)                    ▷ Inference without hidden human context
3:    **for** different temperatures **do**   ▷ Different temperatures are adopted in simulated annealing
4:       **for** $\gamma_1$ iterations **do**
5:          randomly choose layout, apply layout dynamics to optimize layout $V_g^l$
6:       **for** each object $v_i \in V_g^o$ **do**
7:          **for** $\gamma_2$ iterations **do**
8:             randomly apply object dynamics to optimize object $v_i$
9: **procedure** STEP2($V_f^a, \{\tilde{\mu}\}$)                    ▷ Inference of hidden human context
10:    group objects and assign activity labels (see last paragraph in subsection 4.3)
11:    **for** each activity group $v_i \in V_f^a$ **do**
12:       **repeat**
13:          randomly apply human pose dynamics to optimize $\tilde{\mu}_i$
14:       **until** $\mathcal{E}(\tilde{\mu}_i | v_i)$ converges               ▷ Maximizing grouping energy in Equation 11
15: **procedure** STEP3($V_g^o, V_g^l, \{\tilde{\mu}\}$)                    ▷ Iterative inference of whole parse graph
16:    **for** different temperatures **do**
17:       **for** $\gamma_3$ iterations **do**
18:          randomly choose layout, objects or human pose
19:          apply random dynamics to minimize $P(\mathbf{pg}|I)$
20: **Return** $\mathbf{pg}_{\text{optimized}}$

---

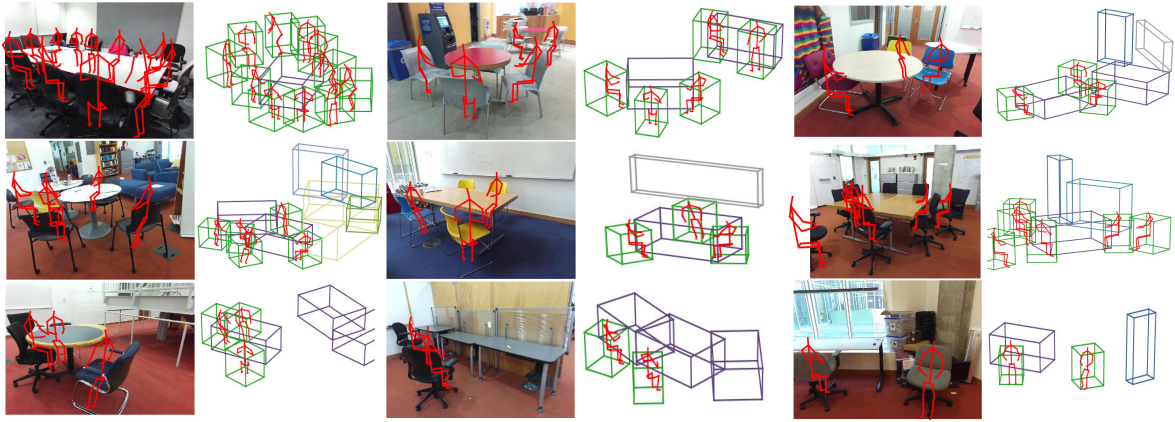

Fig. 4: Sampled human poses in various indoor scenes. Objects in multiple activity groups have multiple poses. We visualize the pose with the highest likelihood.

than 0. For each activity group $v_i \in V_f^a$, the pose of the imagined human is estimated by maximizing the likelihood $p(v_i|\tilde{\mu}_i)$, which is equivalent to minimize the grouping energy $\mathcal{E}_{grp}(\tilde{\mu}_i|v_i)$ defined in Equation 6,

$$y_i^*, m_i^*, t_i^*, r_i^*, s_i^* = \underset{y_i, m_i, t_i, r_i, s_i}{\arg\min} \; \mathcal{E}_{grp}(\tilde{\mu}_i|v_i), \tag{11}$$

Figure 4 shows the results of sampled human poses in various indoor scenes.

## 5  Experiments

We use the SUN RGB-D dataset [45] to evaluate our approach on 3D scene parsing, 3D reconstruction, as well as other 3D scene understanding tasks. The dataset has 5050 testing images and 10,355 images in total. Although it provides RGB-D data, we only use the RGB images as the input for training and testing. Figure 5 shows some qualitative parsing results (top 20%).

Table 1: 3D scene parsing and reconstruction results on SUN RGB-D dataset

| Method | # of image | 3D Layout Estimation IoU | Holistic Scene Understanding | | | |
|---|---|---|---|---|---|---|
| | | | $P_g$ | $R_g$ | $R_r$ | IoU |
| 3DGP [19] | 5050 | 19.2 | 2.1 | 0.7 | 0.6 | 13.9 |
| Ours (init.) | 5050 | 46.7 | 25.9 | 15.5 | 12.2 | 36.6 |
| Ours (joint.) | 5050 | **54.9** | **37.7** | **23.0** | **18.3** | **40.7** |
| 3DGP [19] | 749 | 33.4 | 5.3 | 2.7 | 2.1 | 34.2 |
| IM2CAD [33] | 484 | 62.6 | - | - | - | 49.0 |
| Ours (init.) | 749 | 61.2 | 29.7 | 17.3 | 14.4 | 47.1 |
| Ours (joint.) | 749 | **66.4** | **40.5** | **26.8** | **21.7** | **52.1** |

Table 2: Comparisons of 3D object detection on SUN RGB-D dataset

| Method | bed | chair | sofa | table | desk | toilet | fridge | sink | bathtub | bookshelf | counter | door | dresser | lamp | tv | mAP |
|---|---|---|---|---|---|---|---|---|---|---|---|---|---|---|---|---|
| [19] | 5.62 | 2.31 | 3.24 | 1.23 | - | - | - | - | - | - | - | - | - | - | - | - |
| Ours (init.) | 45.55 | 5.91 | 23.64 | 4.20 | 2.50 | 1.91 | 14.00 | 2.12 | 0.55 | 2.16 | 0.34 | 0.01 | 5.69 | 1.12 | 0.62 | 7.35 |
| Ours (joint.) | **58.29** | **13.56** | **28.37** | **12.12** | **4.79** | **16.50** | **15.18** | **2.18** | **2.84** | **7.04** | **1.6** | **1.56** | **13.71** | **2.41** | **1.04** | **12.07** |

We evaluate our method on three tasks: i) 3D layout estimation, ii) 3D object detection, and iii) holistic scene understanding with all the 5050 testing images of SUN RGB-D across all scene categories. The capability of generalization to all the scene categories is difficult for most of the conventional methods due to the inaccuracy of camera parameter estimation and severe sensitivity to the occlusions in cluttered scenes. In this paper, we alleviate it by using the proposed learning-based camera parameter estimation and a novel method to initialize the geometric attributes. In addition, we also achieve the state-of-the-art results in 2D layout estimation on LSUN dataset [55] and Hedau dataset [28]. The implementation details, and additional results of camera parameter estimation and 2D layout estimation are summarized in the *supplementary material*.

**3D Layout Estimation:** The 3D room layout is optimized using the proposed joint inference. We compare the estimation by our method (with and without joint inference) with 3DGP [19]. Following the evaluation protocol defined in [45], we calculate the average Intersection over Union (IoU) between the free space from the ground truth and the free space estimated by our method. Table 1 shows our method outperforms 3DGP by a large margin. We also improve the performance by 8.2% after jointly inferring the objects and layout, demonstrating the usefulness of integrating the joint inference process.

Since IM2CAD [33] manually selected 484 images from living rooms and bedrooms without releasing the image list, we compare our method with them on the entire set of living rooms and bedrooms. Table 1 shows our method surpasses IM2CAD, especially after incorporating the joint inference process.

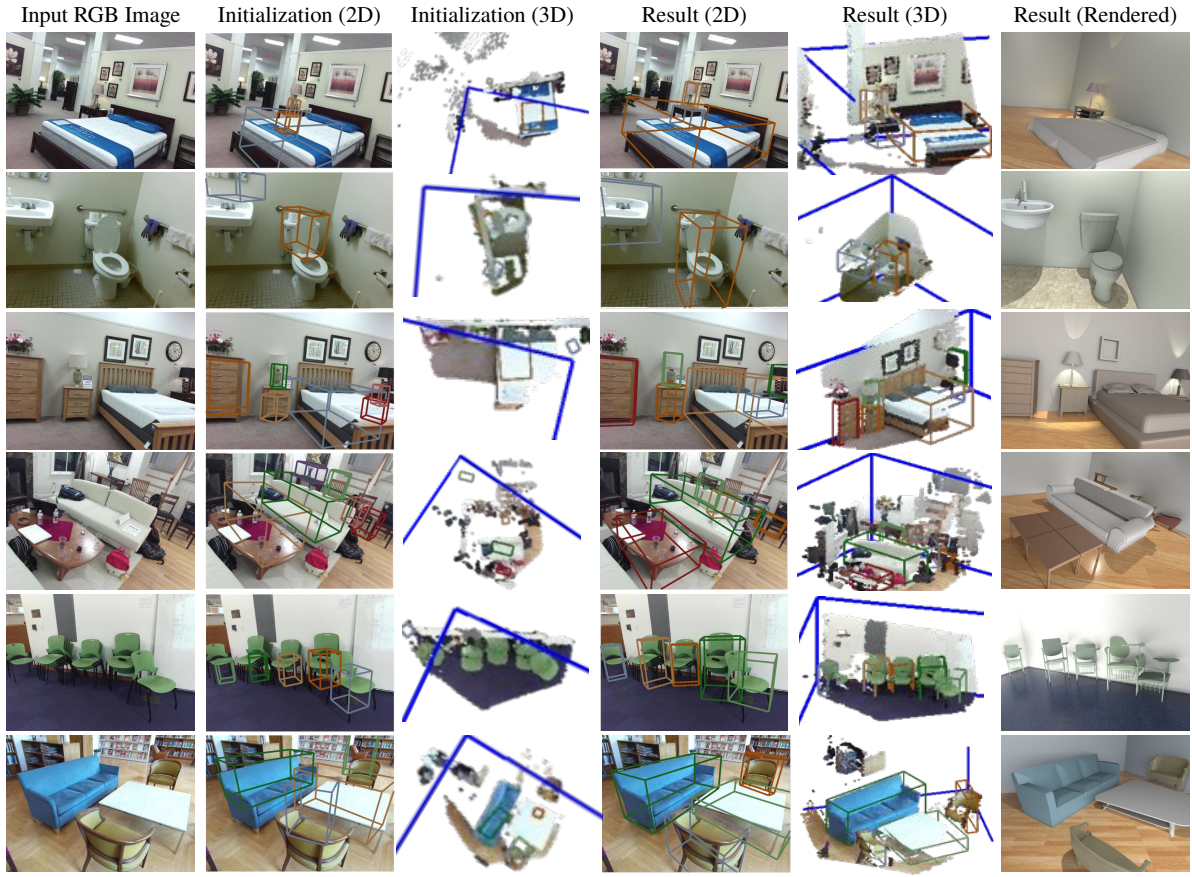

Fig. 5: Qualitative results of the proposed method on SUN RGB-D dataset. The joint inference significantly improves the performance over individual modules.

**3D Object Detection:** We evaluate our 3D object detection results using the metrics defined in [45]. We compute the mean average precision (mAP) using the 3D IoU between the predicted and ground truth 3D bounding boxes. In the absence of depth, we adjust threshold IoU from 0.25 (evaluation setting with depth as the input) to 0.15 and report our results in Table 2. 15 out of 30 object categories are reported here due to the limited space; full table is reported in the *supplementary material*. The results indicate our method not only exceeds the detection score by a significant margin but also makes it possible to evaluate the entire object categories. Note that although IM2CAD also evaluates the detection, they use the metric related to a specified distance threshold. Here, we also compare with IM2CAD on the subset with this special metric rather than IoU threshold. We are able to obtain an mAP of 80.2%, higher than an mAP of 74.6% reported in the IM2CAD.

**Holistic Scene Understanding:** We estimate the detailed 3D scene including both objects and room layout. Using the metrics proposed in [45], we evaluate the geometric precision $P_g$, geometric recall $R_g$, and semantic recall $R_r$ with the IoU threshold set to 0.15. We also evaluate the IoU of the free space (3D voxels inside the room polygon but outside any object bounding box) between the ground truth and the estimation. Table 1 shows that the proposed method demonstrates a significant improvement. Moreover, we improve the initialization result by 12.2% on geometric precision, 7.5% on geometric recall, 6.1% on semantic recall, and 4.1% on free space estimation. The improvement of total scene understanding indicates that the joint inference can largely improve the performance of each task. Using the same setting with 3D layout estimation, we compare with IM2CAD [33] and improve the free space IoU by 3.1%.

Table 3: Ablative analysis of our method on SUN RGB-D dataset. We evaluate on holistic scene understanding under different settings. We denote support relation as $C_1$, physical constraint as $C_2$ and human imagination as $C_3$. Similarly, we denote the setting of only optimizing the layout during inference as $S_4$, only optimizing the objects during inference as $S_5$

| Setting | w/o $C_1$ | w/o $C_2$ | w/o $C_3$ | w/o $(C_1, C_2, C_3)$ | $S_4$ | $S_5$ | All |
|---------|-----------|-----------|-----------|------------------------|-------|-------|-----|
| IoU | 42.3 | 41.3 | 43.8 | 38.4 | 39.4 | 36.3 | **44.7** |
| $P_g$ | 29.3 | 23.5 | 32.1 | 19.4 | 14.9 | 28.4 | **34.4** |
| $R_g$ | 17.4 | 15.6 | 20.4 | 12.4 | 11.2 | 19.7 | **24.1** |
| $R_r$ | 14.1 | 10.5 | 16.5 | 8.7 | 8.6 | 13.3 | **19.2** |

**Ablative Analysis:** The proposed HSG incorporates several key components including supporting relations, physics constraints and latent human contextual relations. To analyze how each component would influence the final results, as well as how much the joint inference process would benefit each task, we conduct the ablative analysis on holistic scene understanding under different settings, through turning on and off certain components or skipping certain steps during joint inference. The experiments are tested on the subset of offices where we incorporate the latent human context. Table 3 summarizes the results. Among all the energy terms we incorporate, physical constraints influence the performance the most, which demonstrates the importance of the physical common sense during inference. It also reflects the efficiency of joint inference as the performances would drop by a large margin without the iterative joint inference.

## 6    Conclusion

We present an analysis-by-synthesis framework to recover the 3D structure of an indoor scene from a single RGB image using a stochastic grammar model integrated with latent human context, geometry and physics. We demonstrate the effectiveness of our algorithm from three perspectives: i) the joint inference algorithm significantly improves results in various individual tasks and ii) outperforms other methods; iii) ablative analysis shows each of module plays an important role in the whole framework. In general, we believe this will be a step towards a unifying framework for the holistic 3D scene understanding.

**Acknowledgments**. We thank Professor Ying Nian Wu from UCLA Statistics Department for helpful discussions. This work is supported by DARPA XAI N66001-17-2-4029, MURI ONR N00014-16-1-2007, SPAWAR N66001-17-2-3602, and ARO W911NF-18-1-0296.

# Holistic 3D Scene Parsing and Reconstruction from a Single RGB Image

# Supplementary Material

Siyuan Huang[1,2], Siyuan Qi[1,2], Yixin Zhu[1,2],
Yinxue Xiao[1], Yuanlu Xu[1,2], and Song-Chun Zhu[1,2]

[1] University of California, Los Angeles
[2] International Center for AI and Robot Autonomy (CARA)

## 1    Learning of Prior Knowledge

The learning process of our method includes two steps: i) collecting the statistics of scene categories, object categories, object sizes, and supporting relations from SUN RGB-D dataset [1]; ii) collecting the statistics of grouping occurrences and the geometric relations between objects and human from Watch-n-Patch [2].

Using SUN RGB-D, we model the prior of scene types, object categories and support relations by multinoulli distributions. For example, a lamp is supported by the floor with a probability of 0.4 and by a desk with a probability of 0.2. The branching probability is simply counting the frequency of each alternative choice. The distribution of the object sizes is learned via non-parametric kernel density estimation.

The human-centric grouping occurrence and human-object interactions in 3D space are learned from the Watch-n-Patch. This dataset collects the RGB-D videos of human activities in offices and kitchens. Since some activities are irrelevant with objects, we learn the activities of 'reading', 'play-computer', 'take-item' and 'put-down-item' in all the office videos. For each activity, we first extract key frames from each sequence with group activity labels. Then we compute the occurrence frequency of the objects around human within a distance threshold, and model the prior of object category using a multinomial distribution. The geometric relations between the objects and humans are similarly learned by fitting normal distributions of relative distance, height, and orientation between each joint of a human pose and the object center.

## 2    2D Room Layout Estimation

Similar to [3], we use a keypoint-based room layout representation to train our network. Figure 1 shows the regular room types defined in [4] with their respective keypoints.

Our model is able to predict both keypoint and room type from an input image using a single model. To achieve this goal, we increase the number of channels in the output layer to match the total number of keypoints (in total 48) of all 11 room types. The cost function is the same as described in [3], which incorporates the Euclidean loss for layout heatmap regression and the cross-entropy loss for room type estimation.

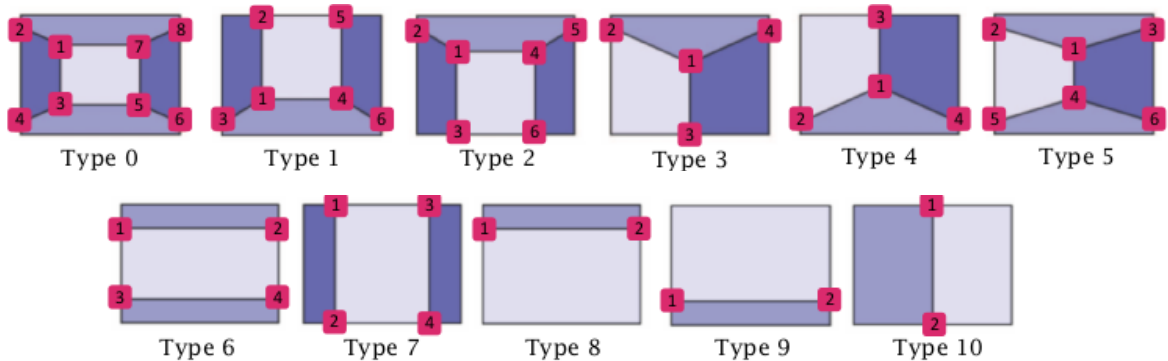

Fig. 1: Types of room layout. The room types are defined in [4]. These 11 room types cover most of the possible configurations of the indoor scenes under Manhattan world assumption [5].

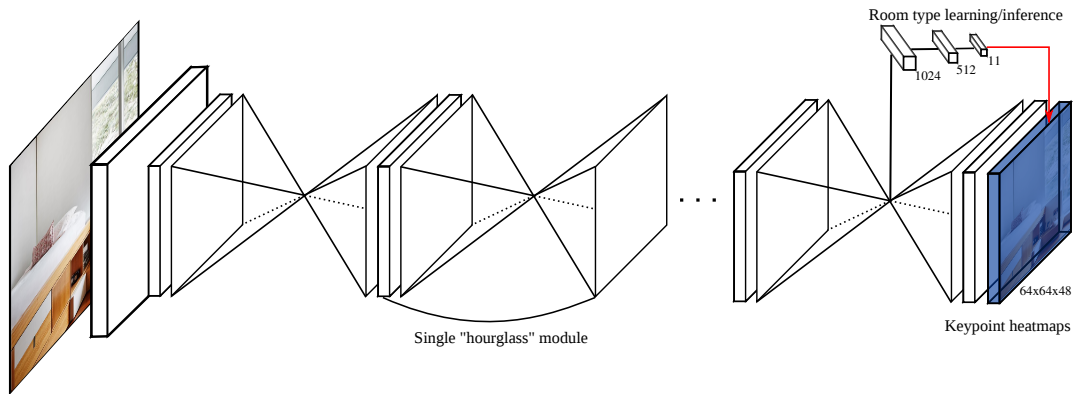

Fig. 2: Network Architecture. The "hourglass" modules work as encoder-decoders which allow for repeated bottom-up, top-down inference.

Figure 2 shows our network architecture. Compared with [3], we use the "stacked hourglass" network [6] as our basic network architecture rather than SegNet [7]. Our network consists of multiple stacked hourglass modules which allow for repeated bottom-up, top-down inference.

The input to the network is 256x256. The output of the network is the room type keypoint heatmaps in a resolution 64x64 within a respect room type category label. We use the Adam optimizer [8] with batch size 16, initial learning rate 0.0001. We train 150 epochs, which takes about 2 days on a 12GB NVIDIA Titan X GPU. We also degrade the gradient of background pixels by multiplying them with a factor of 0.2 to prevent the output converges to zero due to the imbalance between foreground and background distribution.

# 3    Implementation Details

For 2D object detection, we fine-tune the object detector on SUN RGB-D with 30 object categories. Since [4] and [9] have no ground-truth of the camera parameter, we train the 2D layout estimation module using [4] as the initial model, followed by using the feature of the heatmap (stacking three FC layers (512-16-1)) to further train camera parameter and scene category on SUN RGB-D. During the initialization and joint inference process, we use the depth estimation model as described in [10], surface normal estimation in [11], and semantic segmentation in [12]. These models are trained on the training set of the SUN RGB-D or NYU v2 dataset [13] (included in the SUN RGB-D). In this paper, we further incorporate human context inference on the subset of offices and skip it on other scenes. During joint inference, we fix the scene category, object categories and support relations to reduce the computational complexity. We used OpenGL [14] to render the depth, surface normal and segmentation map. Rendering each map takes about 1 second. On average, our joint inference process takes about one hour for each image on a single CPU core.

# 4    Additional Experiment Results

## 4.1    Evaluation of 2D Layout Estimation

We evaluate the 2D layout estimation without joint inference on LSUN dataset [4] and Hedau dataset [9]. The LSUN dataset consists of 4000 training, 394 validation and 1000 test images. The Hedau dataset contains 209 training, 56 validation and 105 test images. We follow the standard evaluation procedure [17] and use pixel errors and keypoint errors as two evaluation metrics. Pixel errors compute the pixel-wise error between the ground truth and estimations of the surface label, and the keypoint errors only considers the average Euclidean distance between the annotated and estimated keypoints. As reported in Table 1, our

Table 1: Quantitative comparisons of 2D layout estimation on LSUN [4] and Hedau dataset [9]

| Method | LSUN | | Hedau |
| --- | --- | --- | --- |
| | Keypoint Error (%) | Pixel Error (%) | Pixel Error(%) |
| Hedau *et al.* (2009) [9] | 15.48 | 24.23 | 21.20 |
| Zhao *et al.* (2013) [15] | - | - | 14.50 |
| Mallya *et al.* (2015) [16] | 11.02 | 16.71 | 12.83 |
| Dasgupta *et al.* (2016) [17] | 8.20 | 10.63 | 9.73 |
| Ren *et al.* (2016) [18] | 7.57 | 5.23 | 8.67 |
| Izadinia *et al.* (2017) [19] | - | 10.04 | 10.15 |
| Lee *et al.* (2017) [3] | 6.30 | 9.86 | 8.34 |
| Zhao *et al.* (2017) [20] | 5.29 | **3.84** | **6.60** |
| Ours (init.) | **5.22** | 4.53 | 7.03 |

4 S. Huang *et al.*

approach achieves 5.22% keypoint error, which outperforms all existing methods and comparable pixel error with the previous best results [20] on both LSUN and Hedau dataset.

### 4.2 Evaluation of Camera Parameter Estimation

We compute the mean absolute error between our estimation and the ground-truth on testing set of SUN RGB-D. As shown in Table 2, comparing with the traditional geometry-based method [9], the proposed method gains a significant improvement. Quantitative results of the comparison over all the scene categories are shown in Figure 3. Empirically, geometry-based methods perform poorly in cluttered scenes (*e.g.*, storage rooms) and perform well in clean scenes with clear orthogonal lines (*e.g.*, receptions). Our method provides a good estimation which applies to most of the indoor scenes, improving the generalization ability of the monocular reconstruction algorithms.

Figure 3 shows the comparison in detail over all categories. We can see that the geometry-based method performs well over the scenes with clear lines in three orthogonal directions like receptions, but results in large errors over cluttered scenes like storage rooms. Our method provides a good estimation which applies to most of the indoor scenes, improve the generalization ability for the *single-view* reconstruction algorithms.

### 4.3 Evaluation of 3D Layout Estimation

Figure 4 shows the comparison with 3DGP over all categories; we can also observe that 3DGP fails in some scene categories such as dinette and cafeteria, which further reflects the drawbacks of the geometry-based methods.

Table 2: Camera parameter estimation.

| Method | Mean Absolute Error | | |
|---|---|---|---|
| | focal length | pitch | roll |
| Hedau et al. [9] | 141.78 | 3.45 | 33.85 |
| Ours | **35.87** | **3.12** | **7.60** |

Table 3: Comparisons of 3D object detection on SUN RGB-D dataset.

| Method | bed | chair | sofa | table | desk | toilet | fridge | sink | bathtub | bookshelf | counter | door | dresser | lamp | tv |
|---|---|---|---|---|---|---|---|---|---|---|---|---|---|---|---|
| [21] | 5.62 | 2.31 | 3.24 | 1.23 | - | - | - | - | - | - | - | - | - | - | - |
| Ours (init.) | 45.55 | 5.91 | 23.64 | 4.20 | 2.50 | 1.91 | 14.00 | 2.12 | 0.55 | 2.16 | 0.34 | 0.01 | 5.69 | 1.12 | 0.62 |
| Ours (joint.) | **58.29** | **13.56** | **28.37** | **12.12** | **4.79** | **16.50** | **15.18** | **2.18** | **2.84** | **7.04** | **1.60** | **1.56** | **13.71** | **2.41** | **1.04** |

| nightstand | books | tvstand | sofachair | cabinet | endtable | dressermirror | person | recyclebin | curtain | whiteboard | mirror | picture | paper | computer |
|---|---|---|---|---|---|---|---|---|---|---|---|---|---|---|
| - | - | - | - | - | - | - | - | - | - | - | - | - | - | - |
| 5.83 | 0.00 | 3.04 | 8.87 | 0.00 | 0.65 | 17.16 | 1.31 | 0.00 | 0.27 | 0.00 | 0.00 | 0.00 | 0.00 | 0.00 |
| **8.80** | **0.02** | **6.69** | **16.99** | **0.48** | **3.15** | **19.43** | **4.04** | **0.63** | **0.40** | **0.20** | **0.00** | **0.00** | **0.00** | **0.00** |

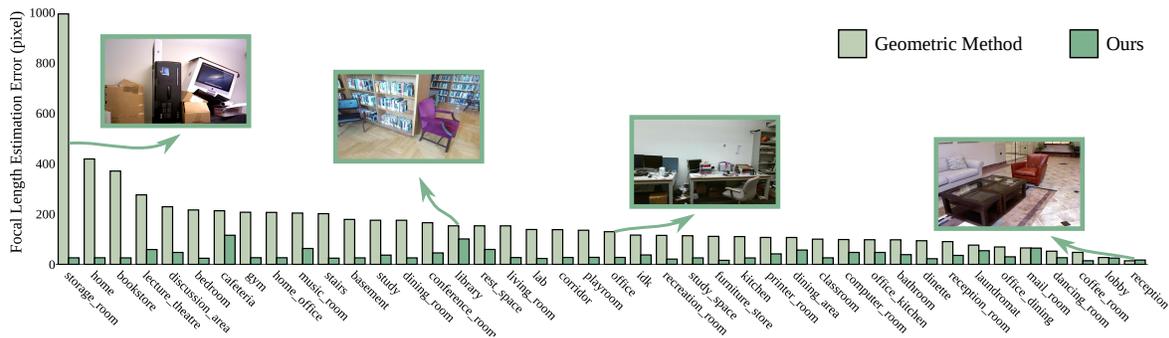

Fig. 3: Estimation error of focal length.

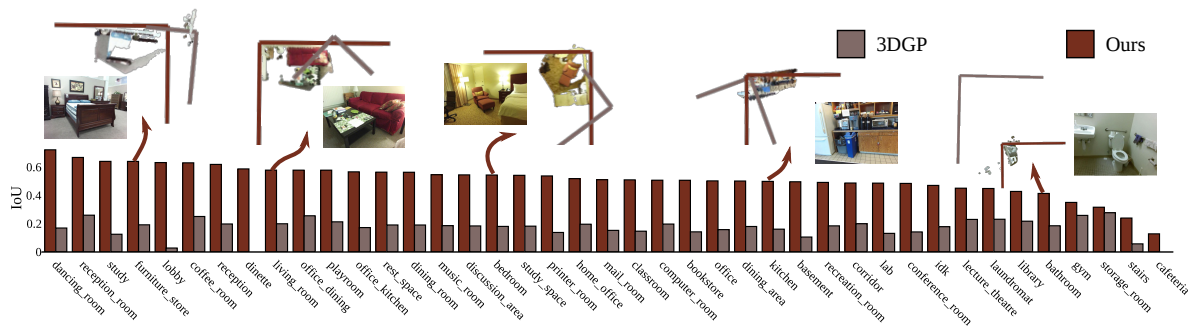

Fig. 4: Quantitative comparisons of 3D layout estimation.

## 4.4   Evaluation of 3D Object Detection

Table 3 shows the evaluation of 3D object detection over 30 categories of objects.

## 5    More Qualitative Results

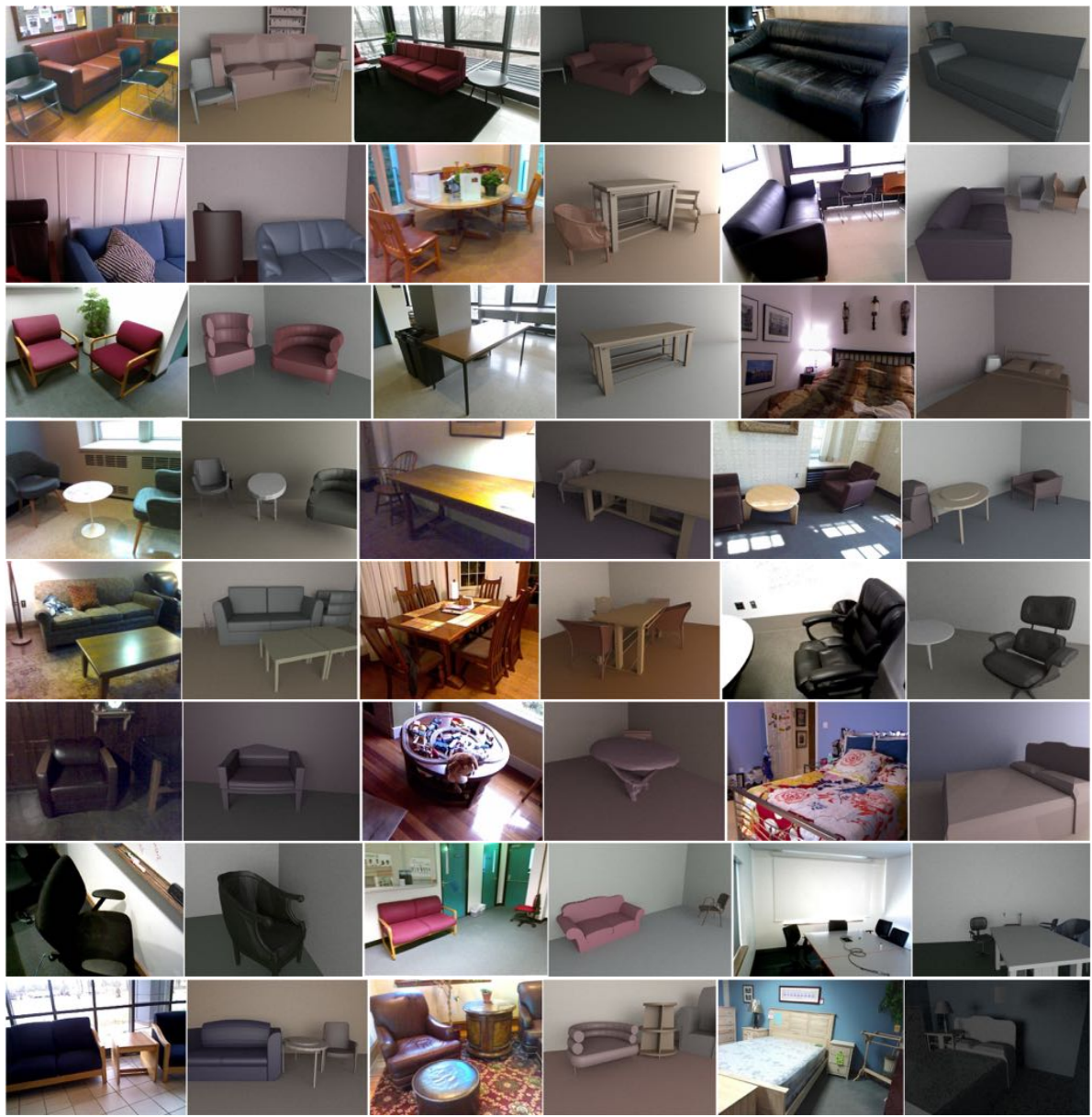

Fig. 5: More qualitative results

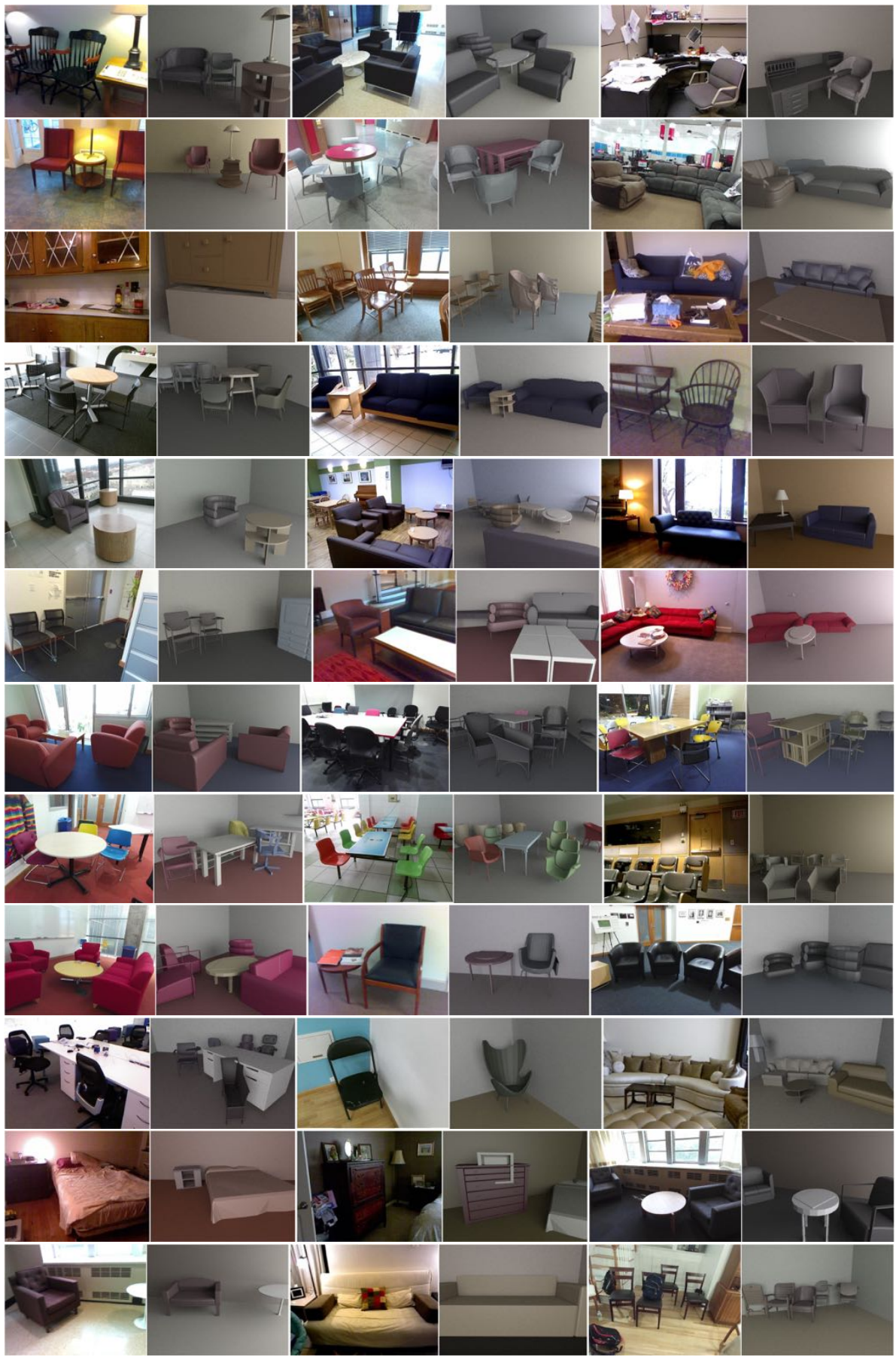

Fig. 5: More qualitative results (cont.)

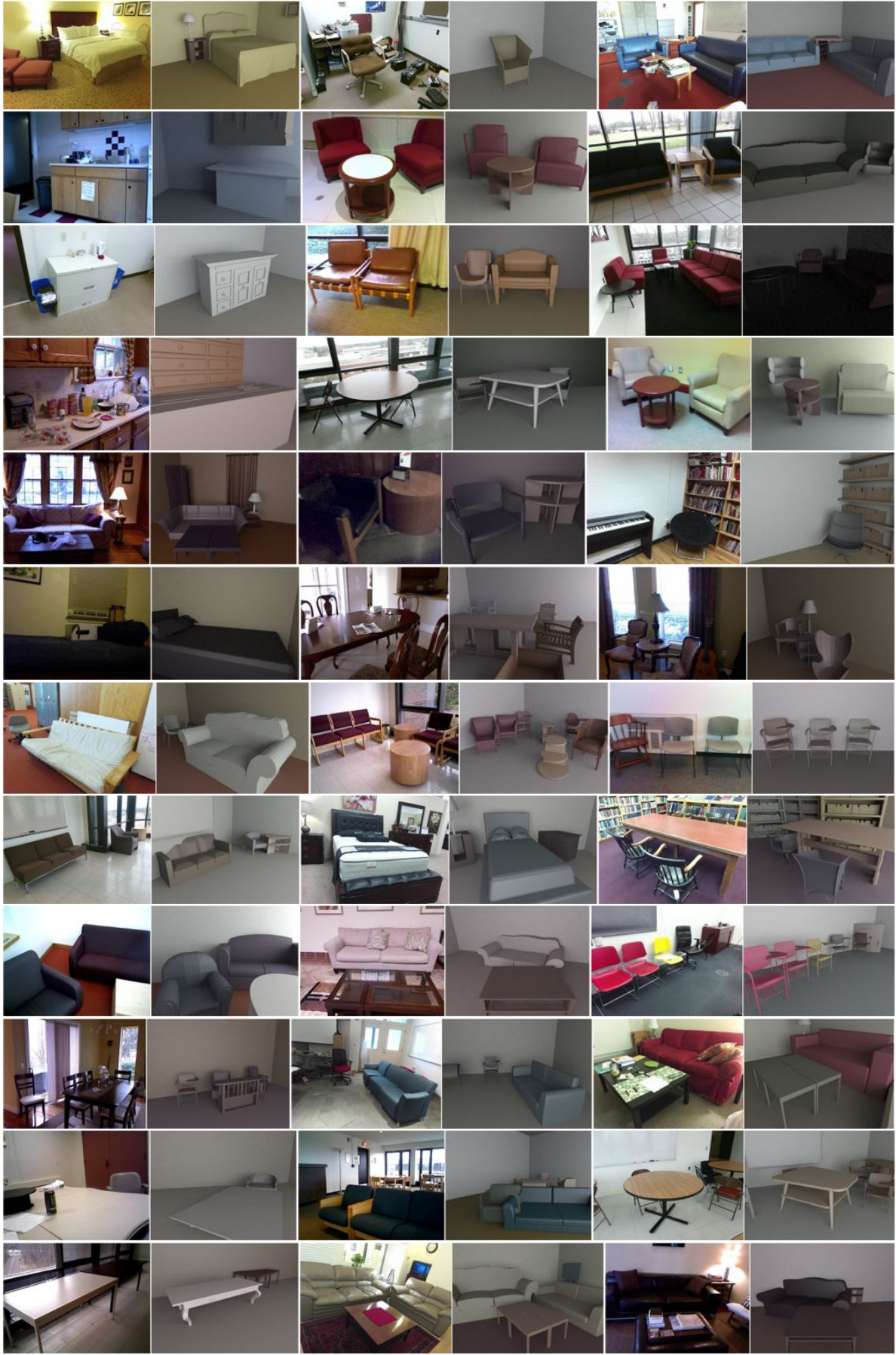

Fig. 5: More qualitative results (cont.)

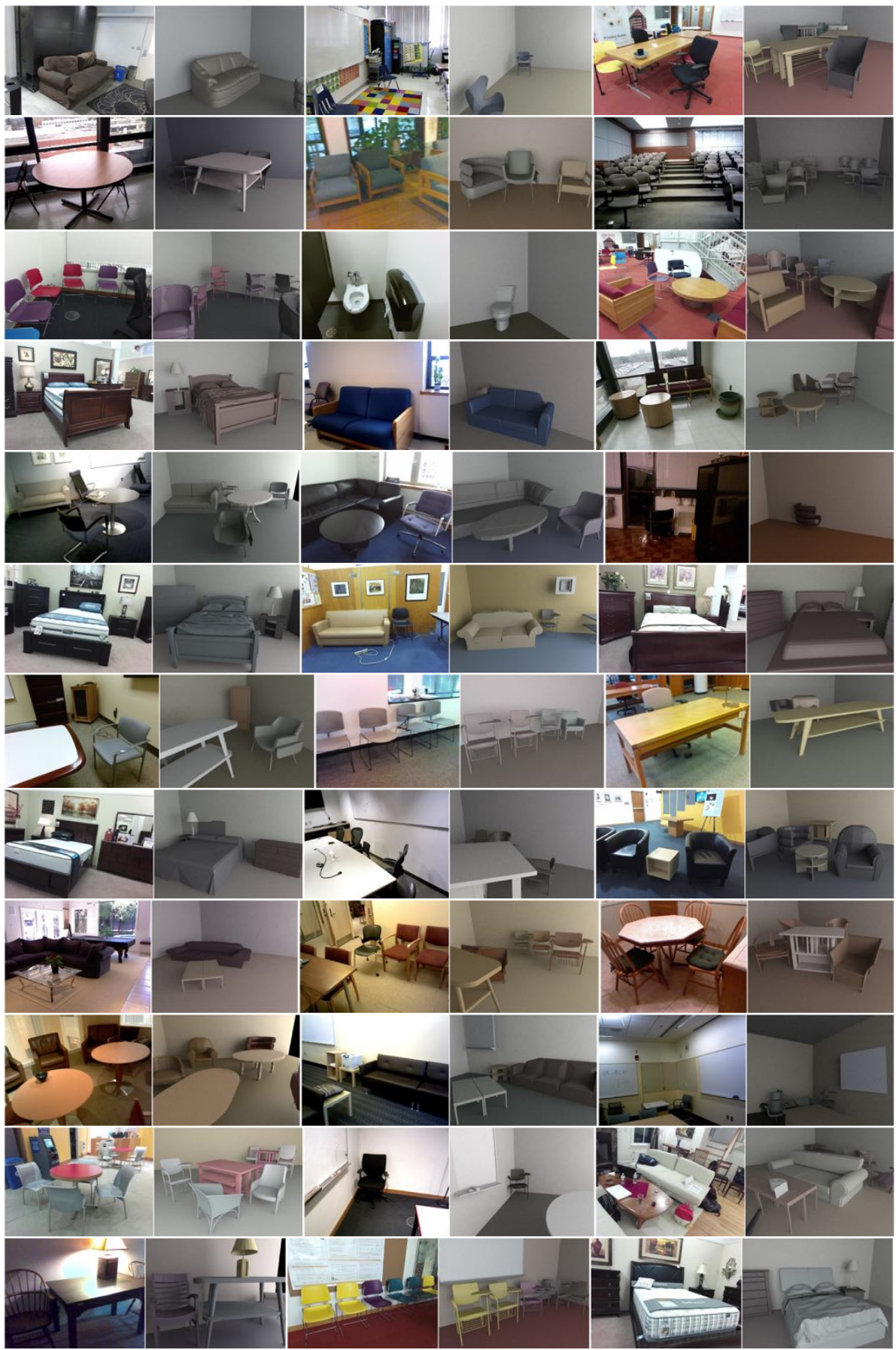

Fig. 5: More qualitative results (cont.)